\documentclass{article}

\usepackage{microtype}
\usepackage{graphicx}
\usepackage{subcaption}
\usepackage{booktabs} 

\usepackage[table]{xcolor}
\usepackage{tabularx}
\usepackage{colortbl}
\usepackage{booktabs}
\usepackage{titlesec}

\definecolor{myblue}{RGB}{240,248,255}
\definecolor{lightgray}{gray}{0.96}

\usepackage{amsmath,amsfonts,bm}
\usepackage{amssymb}

\def\eqref#1{equation~\ref{#1}}

\def\1{\bm{1}}

\DeclareMathAlphabet{\mathsfit}{\encodingdefault}{\sfdefault}{m}{sl}
\SetMathAlphabet{\mathsfit}{bold}{\encodingdefault}{\sfdefault}{bx}{n}

\usepackage[preprint]{arxiv}

\usepackage{amsmath}
\usepackage{amssymb}
\usepackage{mathtools}
\usepackage{amsthm}

\definecolor{citecolor}{HTML}{0071BC} 
\definecolor{linkcolor}{HTML}{ED1C24} 
\usepackage[pagebackref=false, breaklinks=true, letterpaper=true, colorlinks, bookmarks=false,  citecolor=citecolor, linkcolor=linkcolor, urlcolor=gray]{hyperref}

\usepackage[capitalize,noabbrev]{cleveref}

\theoremstyle{plain}

\theoremstyle{definition}

\theoremstyle{remark}

\usepackage[textsize=tiny]{todonotes}

\icmltitlerunning{Pretraining A Large Language Model using Distributed GPUs: A Memory-Efficient Decentralized Paradigm}

\begin{document}

\twocolumn[
  \icmltitle{Pretraining A Large Language Model using Distributed GPUs: \\ A Memory-Efficient Decentralized Paradigm}



  \icmlsetsymbol{corr}{$\dagger$}

  \begin{icmlauthorlist}
    \icmlauthor{Jinrui Zhang}{PolyU,OPPO}
    \icmlauthor{Chaodong Xiao}{PolyU,OPPO}
    \icmlauthor{Aoqi Wu}{PolyU,OPPO}
    \icmlauthor{Xindong Zhang}{OPPO}
    \icmlauthor{Lei Zhang}{PolyU,OPPO}
  \end{icmlauthorlist}

  \icmlaffiliation{PolyU}{Department of Computing, The Hong Kong Polytechnic University.}
  \icmlaffiliation{OPPO}{OPPO Research Institute}

  \icmlcorrespondingauthor{Lei Zhang}{cslzhang@comp.polyu.edu.hk}


  \vskip 0.3in
]



\printAffiliationsAndNotice{}  

\begin{abstract}
  Pretraining large language models (LLMs) typically requires centralized clusters with thousands of high-memory GPUs (\textit{e.g.}, H100/A100). Recent decentralized training methods reduce communication overhead by employing federated optimization; however, they still need to train the entire model on each node, remaining constrained by GPU memory limitations. In this work, we propose \textbf{SP}arse \textbf{E}xpert \textbf{S}ynchronization (\textbf{SPES}), a memory-efficient decentralized framework for pretraining mixture-of-experts (MoE) LLMs. SPES trains only a subset of experts per node, substantially lowering the memory footprint. Each node updates its local experts and periodically synchronizes with other nodes, eliminating full-parameter transmission while ensuring efficient knowledge sharing. To mitigate limited per-expert data utilization under sparse expert updates, we introduce an expert-merging warm-up strategy, where experts exchange knowledge early in training, to rapidly establish foundational capabilities. With SPES, we train a 2B-parameter MoE LLM using 16 standalone 48GB GPUs over internet connections, which achieves competitive performance with centrally trained LLMs under similar computational budgets. We further demonstrate scalability by training a 7B model from scratch and a 9B model upcycled from a dense checkpoint, both of which match prior centralized baselines. Our code is available at \url{https://github.com/zjr2000/SPES}.
  
\end{abstract}

\section{Introduction}

\begin{figure*}[t]
  \centering
  \includegraphics[width=0.78\linewidth]{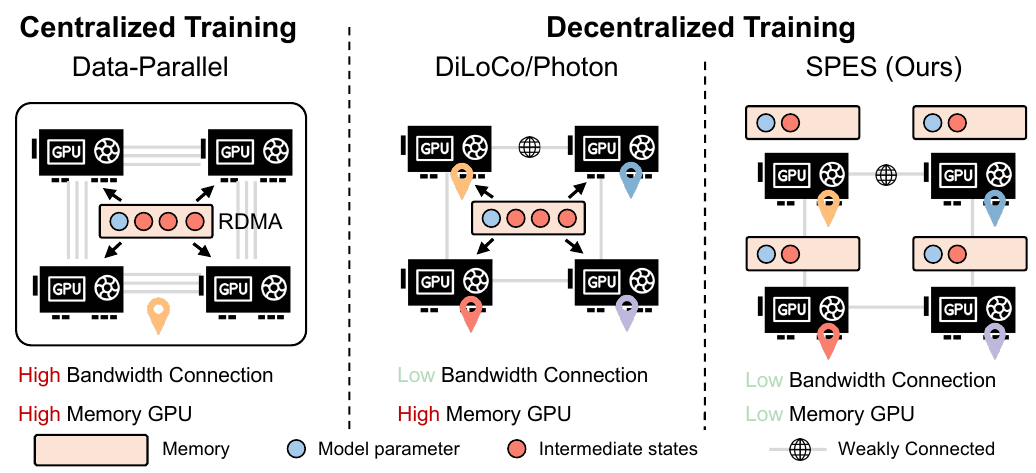}
  \caption{\textbf{Comparison of different pretraining paradigms for LLM.}  \textbf{Left}: centralized training, which requires high-memory GPUs and high-bandwidth interconnects (\textit{e.g.}, RDMA) for its tightly coupled model or data parallelism. \textbf{Middle}: existing decentralized training (\textit{e.g.}, DiLoCo, Photon), where each node trains a full model locally, reducing bandwidth needs but still demanding high-memory GPUs. \textbf{Right}: our proposed SPES, a memory-efficient decentralized method for training MoE-based LLMs, where each node trains only a subset of experts, substantially reducing both per-GPU memory usage and communication overhead.}
  \label{fig:framework_comparison}
\end{figure*}

Large language models (LLMs)~\citep{achiam2023gpt,grattafiori2024llama,yang2025qwen3,liu2024deepseek,muennighoff2024olmoe} have shown strong generalization capabilities across various downstream tasks, establishing themselves as fundamental components in real-world applications such as conversational assistant~\citep{cui2024chatlaw} and embodied agent~\citep{fung2025embodied}. However, pretraining LLMs remains highly resource-intensive. The main bottlenecks arise from the substantial GPU memory required to store model parameters, activations, optimizer states, and gradients, and the need of low-latency, high-bandwidth inter-device communication to support model and data parallelism~\citep{shoeybi2019megatron,rasley2020deepspeed,zhao2023pytorch}. Consequently, existing LLMs are typically trained under centralized settings (as shown in Fig.~\ref{fig:framework_comparison} (left)), utilizing co-located clusters equipped with high-memory GPUs and fast interconnects (\textit{e.g.}, RDMA). For instance, LLaMA3-405B~\citep{grattafiori2024llama} is trained using up to 16K H100 GPUs linked with high-bandwidth interconnects, while OLMo2 7B~\citep{olmo20242} is trained on a cluster of 1,024 H100 GPUs. Such high infrastructure requirements make LLM pretraining inaccessible to most researchers in the community.

To mitigate the demands of centralized LLM training, recent works such as DiLoCo~\citep{douillard2023diloco} and Photon~\citep{sani2024photon} have explored decentralized pre-training paradigms (as shown in Fig.~\ref{fig:framework_comparison} (middle)). In these approaches, each workstation performs local updates and synchronizes with peers intermittently, following a federated optimization protocol (\textit{e.g.,} FedAvg~\citep{mcmahan2017communication}). This sparse communication mode significantly reduces the bandwidth requirements compared to centralized data- or model-parallel methods, enabling training across geographically distributed, heterogeneous GPU clusters. While communication constraints are relaxed, however, these approaches still require each node to update the full set of model parameters. Consequently, the memory footprint per node remains substantial. This limitation is especially significant for training large-scale LLMs, where insufficient memory can be a bottleneck.

To address this challenge, we propose \textbf{SP}arse \textbf{E}xpert \textbf{S}ynchronization (SPES), a memory-efficient, decentralized training paradigm tailored for MoE-based LLMs, as shown in the right panel of Fig.~\ref{fig:framework_comparison}. Compared to dense models, MoE models are inherently well-suited for decentralized environments, as each expert can be managed independently, enabling finer-grained training and resource management.  In SPES, each node is responsible for training a distinct subset of experts, while keeping the remaining experts frozen during local updates. This design substantially reduces the memory requirement, since each node only needs to maintain the gradients and optimizer states for the experts assigned to it \footnote{Note that optimizer states and gradients typically dominate the static memory footprint (excluding activations) in model training. For example, AdamW~\citep{loshchilov2017decoupled} can account for up to 75\% of static memory.}. All nodes periodically synchronize their trained experts with peers, ensuring continuous knowledge sharing across the network. By eliminating the need to transmit the entire model weights, this sparse synchronization approach substantially reduces communication overhead and enables efficient knowledge exchange between nodes.
A challenge in this sparse training regime is the limited token utilization of individual experts. Since each expert is updated using only a subset of the total training tokens, it receives insufficient training signal to develop robust foundational representations. To address this issue, we introduce an expert-merging warm-up strategy: in the early stages of training, we periodically merge each expert with its most similar peers in a weighted average manner, improving the knowledge sharing across experts.

We evaluate the effectiveness of SPES by pretraining MoE LLMs at 2B, 7B, and 9B parameter scales within decentralized settings. Our results show that SPES enables the training of a 2B-parameter MoE LLM on 16 standalone NVIDIA L40S GPUs (48GB) over the internet, achieving performance comparable to centrally trained models under comparable computational budgets. Compared with previous decentralized training frameworks, SPES reduces up to 33.3\% communication cost and significantly lowers per-GPU memory requirements. We further demonstrate the scalability of SPES by training a 7B model from scratch and upcycling a 9B model from a strong dense initialization; both models match the performance of centralized counterparts trained with similar data and compute resources. Ablation studies and in-depth analysis are also provided to validate the design choices of SPES. Our contributions can be summarized as follows: 

\textbf{(i) A memory-efficient decentralized pretraining framework.} We propose SPES, a memory-efficient decentralized framework for pretraining MoE-based LLMs, where each node trains only a subset of experts, significantly reducing per-device memory and communication overhead.

\textbf{(ii) An expert-merging warm-up strategy.} We introduce an expert-merging warm-up strategy to periodically aggregate similar experts during early training, enabling stronger expert representations with sparse decentralized training. 

\textbf{(iii) Empirical validation.} We demonstrate the effectiveness of SPES by training models across multiple scales, utilizing both training from scratch and continual pretraining regimes on weakly connected GPUs. SPES achieves competitive performance on multiple benchmarks, but with \textbf{significantly lower communication and memory costs} compared to previous training paradigms. 

As most decentralized LLM training frameworks are not open-sourced, we implement a custom gRPC-based cross-node synchronization protocol~\citep{grpc} and integrate it into a mainstream LLM pretraining codebase~\citep{muennighoff2024olmoe}. Our model and code are released to facilitate future works on decentralized training. 
\section{Related Work}

\textbf{Decentralized Training.} Decentralized training has been studied for both fine-tuning~\citep{wu2025learning, bai2024federated, sun2024improving} and pretraining~\citep{douillard2023diloco,sani2024photon,jaghouar2024intellect} LLMs. The works on finetuning pretrained LLMs usually target for privacy-preserving adaptation. FATE-LLM~\citep{fan2023fate} explores federated fine-tuning for advertising generation. Subsequent works~\citep{kuang2024federatedscope, zhang2024towards, ye2024openfedllm} extend federated LLM fine-tuning to instruction-tuning settings. To reduce communication and memory costs, parameter-efficient federated fine-tuning methods have been proposed, such as FedLoRA~\citep{yi2023pfedlora} and FedPETuning~\citep{zhang2023fedpetuning}. 

DiLoCo~\citep{douillard2023diloco} and Photon~\citep{sani2024photon} are among the first to study decentralized LLM pretraining. With FedAvg~\citep{mcmahan2017communication}, they achieve comparable perplexities to centrally trained models while substantially reducing communication cost. More recent efforts improve communication efficiency via new optimizers~\citep{iacob2025loc, kolehmainen2025noloco} and architectures tailored to decentralized settings~\citep{douillard2024dipaco}. At larger scale, INTELLECT-1~\citep{jaghouar2024intellect} demonstrates decentralized pretraining of a 10B-parameter model across independent devices, and~\citet{charles2025communication} further validates the scalability of this communication-efficient paradigm. Despite such advances, those methods still incur significant memory and communication overhead due to full-model training and synchronization. In contrast, our SPES only needs to train a subset of parameters per node, substantially reducing both the memory and communication costs; moreover, SPES can be naturally combined with more advanced optimizers and architectures to further improve scalability.

\textbf{Memory-Efficient Pretraining.} Methods to reduce memory in LLM pretraining primarily leverage sharding and parallelism on tightly coupled accelerators. Data parallelism such as ZeRO~\citep{rajbhandari2020zero} and FSDP~\citep{zhao2023pytorch} partition optimizer states, gradients, and model parameters, enabling distributed storage and computation. Model-parallel techniques~\citep{shoeybi2019megatron}—including pipeline, tensor, and expert parallelism—split model computation to accommodate larger architectures. However, these strategies typically assume a centralized cluster with high-bandwidth interconnects to facilitate frequent synchronization. Orthogonal techniques include mixed precision~\citep{micikevicius2017mixed}, activation checkpointing, memory-efficient attention~\citep{dao2022flashattention,dao2023flashattention}, and optimizer quantization~\citep{dettmers20218}. Our proposed SPES enables cross-node expert sharding with sparse synchronization: gradients and optimizer states are distributed across geographically heterogeneous nodes, each of which trains only the MoE experts assigned to it and communicates only necessary updates. SPES is designed for environments with heterogeneous, low-bandwidth interconnects, such as single-GPU nodes where intra-node sharding is infeasible. Moreover, SPES complements existing parallelism paradigms: when multiple GPUs are available per node, SPES can be combined with previous parallelism strategies to maximize memory efficiency and scalability.
\section{Memory-Efficient Decentralized Pretraining}

In this section, we present the details of our proposed \textbf{SP}arse \textbf{E}xpert \textbf{S}ynchronization (\textbf{SPES}), a memory-efficient decentralized pretraining framework for MoE LLMs. SPES partitions expert training across weakly connected nodes and synchronizes weights intermittently, substantially reducing both the memory usage and the communication overhead compared to prior paradigms. We begin with the preliminaries (Section~\ref{sec:method_preliminaries}), followed by the framework overview (Section~\ref{sec:method_arch}), and the methodology details (Section~\ref{sec:method_spes}).

\begin{figure*}[t]
  \centering
  \includegraphics[width=0.88\linewidth]{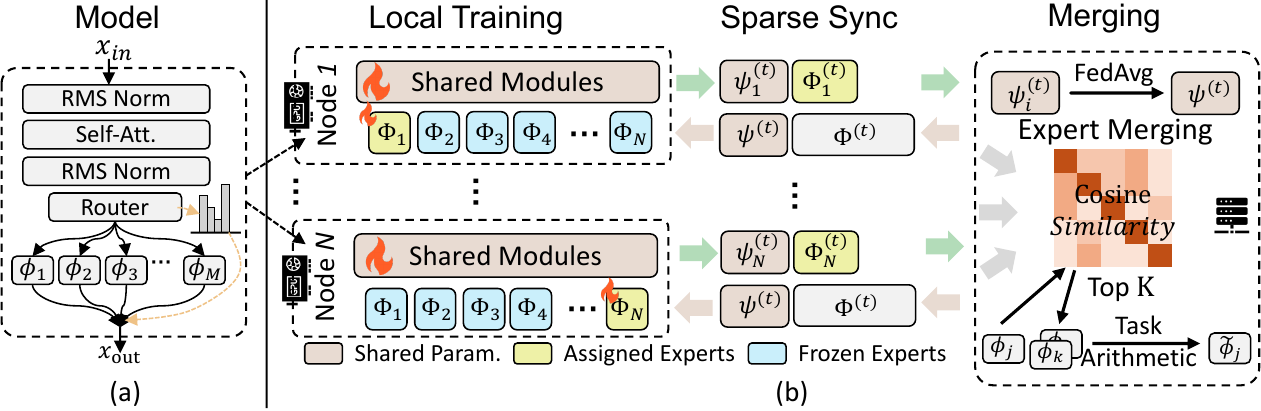}
  \caption{(a) \textbf{Illustration of our model structure}, in which we utilize an MoE LLM comprising standard self-attention blocks, normalization layers, and routed feed-forward modules. (b) \textbf{Illustration of SPES}, where each node performs local training on a disjoint subset of experts to reduce memory consumption. During weight synchronization, only the trained parameters are transmitted, minimizing communication overhead. To improve data utilization, we propose an expert-merging strategy that merges similar experts to facilitate knowledge sharing.}
  \label{fig:spes_framework}
\end{figure*}   

\subsection{Preliminaries}
\label{sec:method_preliminaries}

\textbf{Decentralized Training.} Let $\mathcal{S}=\{\eta_{1},\ldots,\eta_{N}\}$ denote a set of $N$ nodes, where node $\eta_i$ holds local data $\mathcal{D}_i$. Existing decentralized training frameworks~\citep{douillard2023diloco,sani2024photon} often use a two-level optimization scheme: an \emph{outer} optimizer that coordinates global synchronization and an \emph{inner} optimizer that performs local updates. In the $t^{th}$ communication round, the global parameters obtained in the previous round, denoted by $\bm{\theta}^{\smash{(t-1)}}$, are broadcast to all nodes. Each node runs $H$ steps the inner optimizer (\textit{e.g.}, AdamW~\citep{loshchilov2017decoupled}) on its shard $\mathcal{D}_i$, producing the updated local parameters $\bm{\theta}_i^{\smash{(t)}}$. The global model aggregates local updates by averaging the model deltas, updating its parameters via
\begin{equation}
\label{equ:outer_opt}
\bm{\theta}^{(t)} \leftarrow \mathrm{OuterOpt}\Big(\bm{\theta}^{(t-1)}, \tfrac{1}{N} \textstyle\sum_{i=1}^N (\bm{\theta}_i^{(t)} - \bm{\theta}^{(t-1)})\Big).
\end{equation}
When the outer optimizer is set to SGD, the above training procedures become the FedAvg~\citep{mcmahan2017communication}, which enables distributed training while minimizing communication overhead. However, each node is required to train the entire model, which needs to store a large amount of intermediate optimizer states, limiting its applicability to memory-constrained devices.

\textbf{Mixture-of-Experts LLM.} MoE architectures~\citep{lepikhin2020gshard, muennighoff2024olmoe, dai2024deepseekmoe} extend transformer LLMs by introducing a set of $M$ expert sub-networks $\{\mathcal{E}_j\}_{j=1}^M$, each sub-network $\mathcal{E}_j$ being parameterized by $\bm{\phi}_j$. Given an input token $x$, a gating function $\mathcal{G}(x)$ is used to select a sparse subset of experts to process it. The output of the MoE block is computed as a weighted sum of the selected experts:
\begin{equation}
\mathrm{MoE}(x)=\textstyle\sum_{j=1}^{M}\mathcal{G}_j(x)\,\mathcal{E}_j(x).
\end{equation}
where $\mathcal{G}_j(x)$ denotes the gate weight for expert $j$. By activating only a few experts per token, MoE scales model capacity without a proportional increase in per-token computation.

\subsection{Overall Framework}
\label{sec:method_arch}
Previous sharding strategies, such as FSDP~\citep{zhao2023pytorch} and ZeRO~\citep{rajbhandari2020zero}, partition LLM model training in centralized data-parallel setups. Each node is responsible for a subset of model modules, which alleviates individual memory constraints. However, when inter-node communication bandwidth is limited, the tight coupling between model shards may lead to suboptimal performance due to insufficient synchronization of model updates. To address this issue, we adopt the MoE architecture to train the LLM, where expert modules can be managed independently, thus relaxing synchronization requirements and enabling fine-grained resource allocation. Following prior works~\citep{touvron2023llama, olmo20242, bai2023qwen}, we employ a standard decoder-only MoE LLM, which is composed of self-attention layers, sparse expert feed-forward networks selected via softmax routing, and normalization layers, as illustrated in Fig.~\ref{fig:spes_framework}(a). Positional encoding is implemented using RoPE~\citep{su2024roformer}, SwiGLU~\citep{shazeer2020glu} is adopted as the activation function, and normalization is performed with RMSNorm~\citep{zhang2019root}. QK-Norm is applied to enhance stability. Specifically, we utilize the drop-less MoE~\citep{gale2023megablocks}, as suggested by~\citet{muennighoff2024olmoe}, to maximize expert utilization. 

In this work, our goal is to train an MoE-based LLM using distributed GPUs. Compared to traditional centralized training, the key challenge of our decentralized training lies in the memory and communication bottlenecks. We therefore propose Sparse Expert Synchronization (SPES) to solve this issue. As illustrated in Fig.~\ref{fig:spes_framework}(b), we take advantage of the inherent modularity of MoE LLM by distributing expert training across the
$N$ nodes. Each node is assigned with some shared modules and a unique subset of the $M$ experts, allowing memory-efficient local updates. During training, the nodes perform efficient synchronization to share knowledge. To improve data utilization for each expert, we further propose an expert-merging warm-up strategy. The details of our SPES are presented in the following section.

\subsection{Sparse Expert Synchronization}
\label{sec:method_spes}
\textbf{Expert Assignment and Local Training.}  We denote by $\bm{\Phi} = \{\bm{\phi}_j\}_{j=1}^M$ the set of parameters of all experts. Refer to Fig.~\ref{fig:spes_framework}(b), we partition $\bm{\Phi}$ into $N$ disjoint subsets, so that $\bm{\Phi} = \bm{\Phi}_1 \cup \bm{\Phi}_2 \cup \ldots \cup \bm{\Phi}_N$, where $\bm{\Phi}_i$ denotes the subset of experts assigned to node $\eta_i$. We denote by $\smash{\overline{\bm{\Phi}}}_{i}$ the set of unassigned experts for node $\eta_i$, and denote by $\bm{\psi}_i$ the parameters of the shared modules. At the start of each local training round $t$, node $\eta_i$ receives the global model parameters updated at round $t-1$ and then performs $H$ rounds of local updates on its local data $\mathcal{D}_i$. The designated expert parameters $\bm{\Phi}_i$ and the shared parameters $\bm{\psi}_i$ will be optimized while keeping $\smash{\overline{\bm{\Phi}}}_{i}$ fixed. The updated local parameters at round $t$ can be denoted as:
\begin{equation}
\bm{\theta}_i^{(t)} = \left(\bm{\psi}_i^{(t)},\, \bm{\Phi}_i^{(t)},\, \overline{\bm{\Phi}}_i^{(t-1)} \right).
\end{equation}
Although each node stores a full copy of the model, gradients and optimizer states are maintained only for the updated parameters, which substantially reduces memory overhead.

\textbf{Sparse Synchronization.}
At the end of each local training round $t$, node $\eta_i$ holds updated local parameters $\bm{\theta}_i^{\smash{(t)}}$, where the shared parameter $\bm{\psi}_i$ and the assigned experts $\bm{\Phi}_{i}$ are updated. During synchronization, each node transmits the updated parameters. Shared parameters are aggregated using FedAvg~\citep{mcmahan2017communication}, while experts are updated via direct assignment:
\begin{equation}
\bm{\theta}^{(t)} = \left(
\textstyle
    \frac{1}{N}\sum_{i=1}^N \bm{\psi}_{i}^{(t)} ,\,
    \bigcup_{i=1}^{N} \bm{\Phi}_{i}^{(t)}
\right).
\end{equation}
The aggregated global parameters $\bm{\theta}^{(t)}$ are then broadcast to all nodes for the next round of training. By synchronizing only assigned experts and shared parameters, SPES substantially reduces communication overhead, enabling scalable decentralized training under limited bandwidth.

\textbf{Expert-Merging Warm-Up.} While achieving notable memory efficiency, SPES faces a practical challenge in sparse training: each node updates only its local experts, leaving many tokens assigned to frozen (unassigned) experts without contributing to gradient updates. This leads to lower token utilization compared to centralized training with an equivalent token budget. To address this issue, we propose an expert-merging warm-up strategy to improve token utilization. The core idea is to periodically merge parameters of similar experts across nodes during synchronization. 

Instead of updating each expert solely with local assignments, we identify peer experts with similar input projections and merge their parameters to facilitate knowledge sharing. Specifically, for the $j$-th expert, we compute pairwise cosine similarities between input projection layers:
\begin{equation}
A_{j,k} = \frac{\langle \bm{w}_j^{\mathrm{in}},\ \bm{w}_k^{\mathrm{in}} \rangle}{\| \bm{w}_j^{\mathrm{in}} \|_2\, \| \bm{w}_k^{\mathrm{in}} \|_2},\quad j, k \in \{1,\ldots,M\}, 
\end{equation}
where $\smash{\bm{w}_j^{\mathrm{in}}}$ denotes the input projection weights of expert $\mathcal{E}_j$, for which we select the $K$ most similar experts $\mathcal{Q}_j = \mathrm{TopK}_k(A_{j,k})$, excluding itself. We then update $\mathcal{E}_j$ via task arithmetic~\citep{ilharco2022editing}:
\begin{equation}
\textstyle
\widetilde{\bm{\phi}}_j^{(t)} = \bm{\phi}_j^{(t)} + \alpha\,\frac{1}{K}
\sum_{k \in \mathcal{Q}_j} \bigl( \bm{\phi}_k^{(t)} - \bm{\phi}_j^{(t)} \bigr),
\end{equation}
where $\alpha$ controls the merge strength. To preserve expert specialization during later training stages, we perform merging only within the initial $T_{\mathrm{w}}$ steps, using an interval of $T_{\mathrm{i}}$, and linearly decay $\alpha$ to zero. This strategy enables each expert to benefit from gradients from multiple nodes, which improves token utilization and accelerates knowledge acquisition in decentralized sparse training settings.

\textit{We also provide a theoretical convergence analysis of SPES; please refer to \textbf{Appendix~\ref{sec:theory}} for details.}

\begin{figure*}[!t]
  \centering
  \includegraphics[width=\linewidth]{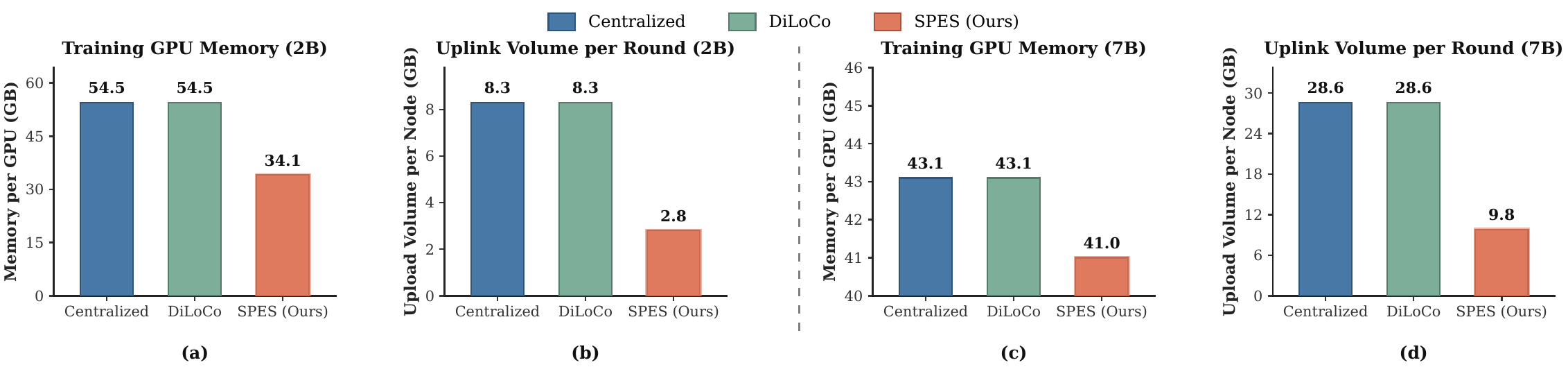}
  \caption{\textbf{Memory and communication costs across training paradigms.} Experiments are conducted with a batch size of 2 and a sequence length of 2048. For the 2B model, we employ PyTorch DDP. For the 7B model, we utilize FSDP across 8 GPUs.}
  \label{fig:memory_and_comunication_comparison}
\end{figure*}

\textbf{Efficiency Analysis.} SPES achieves substantial improvements in both memory and communication efficiency compared to conventional decentralized training methods. For example, when using the AdamW, DiLoCo~\citep{douillard2023diloco} requires each node to store optimizer states and gradients for all model parameters, resulting in a memory cost of $4 \times (|\bm{\psi}| + |\bm{\Phi}|)$ and a communication cost of $2 \times N \times (|\bm{\psi}| +|\bm{\Phi}|)$ per round. In contrast, SPES exploits expert partitioning, and each node only needs to store the intermediate states for the shared parameters and the assigned experts, which reduces the per-node memory cost to $4 \times |\bm{\psi}| + |\bm{\Phi}| + 3 \times |\bm{\Phi}_{i}|$. Similarly, communication overhead is also significantly reduced, as only shared parameters and updated experts are synchronized, resulting in a cost of $N \times (2 \times |\bm{\psi}| + |\bm{\Phi}| + |\bm{\Phi}_{i}|)$ per round. SPES achieves significant reductions in both memory and communication cost, especially as the number of nodes increases. For instance, when training a 2B-parameter MoE model with 16 experts in 16 nodes (one GPU per node; see Fig.~\ref{fig:memory_and_comunication_comparison} for details), DiLoCo requires 55GB of memory per node, whereas SPES reduces this requirement to 35GB. In addition, SPES achieves a 33.3\% reduction in communication cost.

\textbf{Training Losses.} Our model is trained with three losses: standard cross-entropy loss for next token prediction, z-loss~\citep{chowdhery2023palm,zoph2022st} for enhancing training stability, and a load-balancing loss~\citep{lepikhin2020gshard} to encourage uniform expert utilization. Within each node, PyTorch FSDP and mixed-precision are used to further improve memory efficiency. For cross-node synchronization, we use our customized gRPC-based communication protocol.

\section{Experiments}

\begin{table*}[t]
\caption{\textbf{Performance comparison across different training paradigms.} For a controlled comparison, all methods are trained with the same 1B-parameter MoE architecture and 50B training tokens using the SlimPajama dataset.}
\centering
\small
\begin{tabular}{lccccccccc}
\toprule
Method & ARC(e) & ARC(c) & PIQA & SciQ & OBQA & BoolQ & SIQA & WinoGrande & Avg. \\
\midrule
Centralized & 50.18 & 25.08 & \textbf{67.08} & \textbf{77.50} & 31.40 & 55.50 & 42.48 & \textbf{50.91} & 50.02 \\
DiLoCo      & 46.84 & 25.08 & 67.03 & 76.90 & 30.00 & 60.09 & 42.22 & 50.12 & 49.79 \\
\rowcolor{myblue}
SPES        & \textbf{52.11} & \textbf{26.42} & 66.27 & 76.50 & \textbf{31.80} & \textbf{60.95} & \textbf{42.73} & 50.83 & \textbf{50.95} \\
\bottomrule
\end{tabular}
\label{tab:performance_comparison_3_frameworks}
\end{table*}

\begin{table}[t]
\caption{\textbf{Model configurations.} We report the number of activated versus total parameters (\#Param), layers (\#L), attention heads (\#H), intermediate size (Interm.), total experts (\#Exp.), and activated experts per token (\#Act.).}
\label{tab:model-configs}
\centering 
\small
\setlength{\tabcolsep}{5pt} 
\begin{tabular}{lcccccc}
\toprule
\textbf{\#Param} & \textbf{\#L} & \textbf{\#H} & \textbf{Hidden} & \textbf{Interm.} & \textbf{\#Exp.} & \textbf{\#Act.} \\
\midrule
0.3B/1.1B & 12 & 12 & 768  & 2048 & 16 & 2 \\
0.8B/2.1B & 16 & 24 & 1536 & 1280 & 16 & 2 \\
1.6B/7.3B & 16 & 16 & 2048 & 2048 & 32 & 4 \\
3.1B/9.4B & 28 & 16 & 2048 & 6144 & 8 & 2 \\
\bottomrule
\end{tabular}

\end{table}

\begin{figure*}[t]
  \centering
  \includegraphics[width=\linewidth]{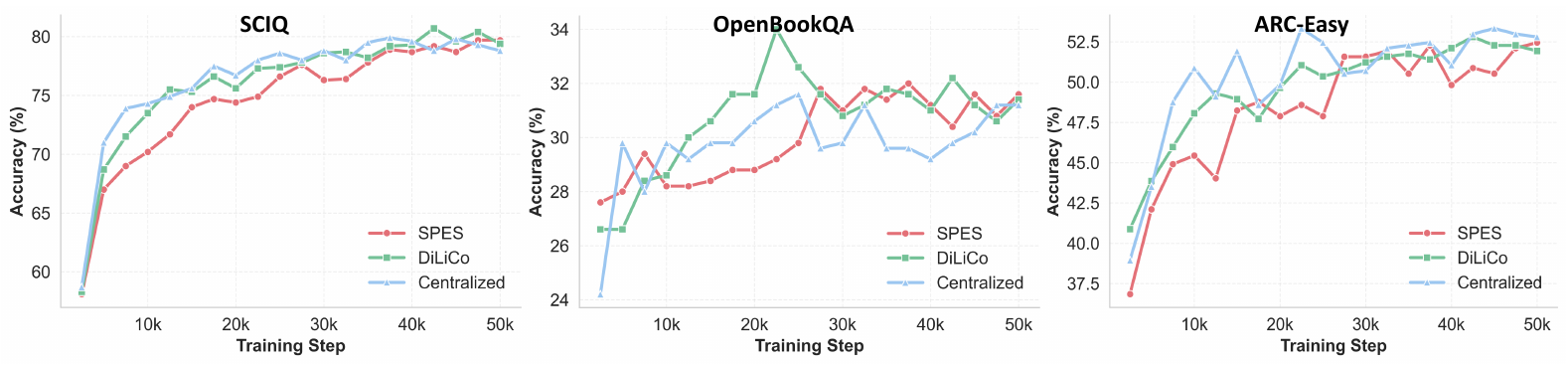}
  \caption{\textbf{Performance comparison across different training paradigms.} Performance during training is evaluated using the evaluation suite integrated into the open-source OLMo codebase.}
  \label{fig:result_comparison_3_frameworks}
\end{figure*}

\begin{table*}[!t]
  \caption{\textbf{Performance comparison with previous LLMs}. $^{*}$ denotes models initialized from the pretrained dense model.}
  \label{tab:commonsense}
  
  \small
  \renewcommand{\arraystretch}{1.1}
  \centering
  \begin{tabular}{
            l    
            >{\centering\arraybackslash}p{1.2cm}  
            >{\centering\arraybackslash}p{1.1cm}  
            >{\centering\arraybackslash}p{0.7cm}  
            >{\centering\arraybackslash}p{0.7cm}  
            >{\centering\arraybackslash}p{0.7cm}  
            >{\centering\arraybackslash}p{0.8cm}  
            >{\centering\arraybackslash}p{1.0cm}  
            >{\centering\arraybackslash}p{1.0cm}  
          }
    \toprule
    \textbf{Method} & \textbf{\#Params} & \textbf{\#Tokens} & \textbf{SciQ} & \textbf{PIQA} & \textbf{SIQA} & \textbf{BoolQ} & \textbf{ARC(e)} & \textbf{ARC(c)} \\
    \midrule
    \multicolumn{9}{c}{\textbf{Models Trained with Significantly More Tokens}} \\
    \midrule
    \rowcolor{lightgray}
    Qwen2.5-0.5B~\citep{qwen2025qwen25technicalreport} & 0.5B/0.5B & 18T & 93.0 & 69.9 & 47.1 & 61.7 & 64.6 & 35.8 \\
    \rowcolor{lightgray}
    Qwen3-0.6B~\citep{yang2025qwen3} & 0.6B/0.6B & 36T & 93.5 & 70.1 & 46.9 & 69.7 & 65.5 & 45.9 \\
     \rowcolor{lightgray}
    Llama3.2-1B~\citep{dubey2024llama} & 1.1B/1.1B & 9T & 91.3 & 73.7 & 45.0 & 63.7 & 71.6 & 43.5   \\
    \rowcolor{lightgray}
    Qwen2.5-1.5B~\citep{qwen2025qwen25technicalreport} & 1.5B/1.5B & 18T & 94.1 & 75.8 & 53.5 & 72.6 & 75.3 & 53.9 \\
    \rowcolor{lightgray}
    SmolLM2-1.7B~\citep{allal2025smollm2} & 1.7B/1.7B & 11T & 93.2 & 77.4 & 46.7 & 72.4 & 77.8 & 54.1 \\
    \rowcolor{lightgray}
    Qwen3-1.7B~\citep{yang2025qwen3} & 1.7B/1.7B & 36T & 95.6 & 75.6 & 52.2 & 79.3 & 73.7 & 55.1 \\
    \rowcolor{lightgray}
    OLMoE-1B-7B~\citep{muennighoff2024olmoe} & 1.3B/7B & 5T & 94.9 & 80.6 & 47.8 & 74.4 & 78.0 & 55.2 \\
    
    \midrule
    \multicolumn{9}{c}{\textbf{Models with $\leq 3$B Parameters}} \\
    \midrule
    OpenELM-0.5B~\citep{mehta2024openelm} & 0.5B/0.5B & 1.5T & 87.2 & 72.3 & - & 55.8 & 48.1 & 27.6 \\
    MobiLlama-0.8B~\citep{thawakar2024mobillama} & 0.8B/0.8B & 1.3T & 85.9 & 73.2 & 43.1 & 60.0 & 49.6 & 28.8 \\
    TinyLlama-1.1B~\citep{zhang2024tinyllama} & 1.1B/1.1B & 3T & 88.9 & 73.3 & - & 57.8 & 55.3 & 30.1 \\
    OpenELM-1.1B~\citep{mehta2024openelm} & 1.1B/1.1B & 1.5T & 90.6 & 75.6 & - & 63.6 & 55.4 & 32.3   \\
    OPT-1.3B~\citep{zhang2022opt}  & 1.3B/1.3B & 180B & 84.3 & 71.7 & 43.7  & 57.7 & 57.0 & 29.7\\
    MobiLlama-1.3B~\citep{thawakar2024mobillama} & 1.3B/1.3B & 1.3T & 89.1 & 74.8 & 44.7 & 60.3 & 56.7 & 36.7  \\
    Pythia-1.4B~\citep{biderman2023pythia}   & 1.4B/1.4B & 300B & 86.4 & 70.9 & 44.6  & 63.3  & 60.7 & 31.2 \\
    OPT-2.7B~\citep{zhang2022opt} & 2.7B/2.7B & 180B & 85.8 & 73.1 & 44.1  & 60.4  & 60.8  & 34.0 \\
    Pythia-2.8B~\citep{biderman2023pythia}  & 2.8B/2.8B & 300B & 88.3 & 74.0 & 44.5 & 64.7 & 66.4 & 36.4 \\
    Open-LLaMA-3B~\citep{openlm2023openllama} & 3B/3B & 1T & 91.8 & 76.2 & - & - & 66.5 & 39.0 \\
    
    \rowcolor{myblue}
    \textbf{SPES-2B (ours)} & 0.8B/2.1B & 500B & 85.0 & 69.3 & 42.3 & 61.4 & 63.8 & 35.3 \\

    \midrule    
    \multicolumn{9}{c}{\textbf{Models with $\geq$ 7B Parameters}} \\
    \midrule
    MoE++ 7B~\citep{jin2024moe++} & 1.2B/7B  & 1T & 89.7 & 78.0 & 45.7 & 64.9  & 66.9 & 43.2 \\
    LLaMA-MoE-3.0B~\citep{llama-moe}$^{*}$ & 3.0B/7B & 2.2T & 89.9 & 77.5 & - & - & 66.8 & 40.9 \\
    OpenMoE-8B/32E~\citep{xue2024openmoe} & 2.1B/8B & 1.1T & - & 74.2 & - & 61.2 & 64.1 & 30.3 \\
    \rowcolor{myblue}
    \textbf{SPES-7B (ours)} & 1.6B/7B & 500B & 89.9 & 74.7 & 44.8 & 62.7 & 72.1 & 43.8 \\
    \rowcolor{myblue}
    \textbf{SPES-9B (ours)$^{*}$} & 3.1B/9B & 400B & 95.3 & 78.9 & 47.5 & 77.3 & 81.5 & 57.3 \\
    \bottomrule
  \end{tabular}
\end{table*}

\subsection{Experiments Setup}

\label{exp:setup}
\textbf{Implementation Details.}
Under training-from-scratch settings, we conduct experiments by training our SPES models at three scales: 1B, 2B, and 7B parameters (see Table~\ref{tab:model-configs} for detailed configurations). All ablation studies are performed on the 1B model, while the 2B and 7B models are trained to compare with previous work. For the 7B model, our training is distributed over $N=4$ compute nodes, each equipped with 8 NVIDIA A800 GPUs interconnected via NVLink, a 96-core Intel Xeon processor (2.90 GHz), and 1.44TB RAM. The nodes communicate over a 13 Gbps Ethernet network, with each node training eight experts (approximately 2.5B trainable parameters per node). For the 2B model, training is performed on $N=16$ nodes, each hosting one NVIDIA L40S GPU, a 64-core Intel Xeon Gold 6148 processor (2.40 GHz), and 720GB RAM. The nodes are connected via 17 Gbps Ethernet. Each node trains one expert, with roughly 0.7B trainable parameters.

Under upcycling settings, we train a 9B model initialized from Qwen3-1.7B-Base~\citep{yang2025qwen3}. We expand the model by replicating the FFN × 8, then inject Gaussian noise into 50\% of parameters with standard deviation 0.02, following~\citet{team2024qwen2}. To match the output scale of the pretrained dense model, we normalize the gating scores after top-k expert selection, following~\citet{jiang2024mixtral}. 

In our implementation, the expert merging warmup steps, $T_{\text{w}}$, is set to 12,500, and the merging interval, $T_{\text{i}}$ is set to 300. The parameters $\alpha$ and $K$ are set to 0.1 and 4, respectively. All models are trained with the AdamW optimizer~\citep{loshchilov2017decoupled}. Please refer to \textbf{Appendix~\ref{appendix:implementation_details}} for additional implementation details.

\textbf{Training Data.}
We train our models exclusively on publicly available datasets, ensuring accessibility for the research community. The 2B and 7B models are trained on data sampled from Ultra-FineWeb~\citep{wang2025ultra} and SlimPajama~\citep{cerebras2023slimpajama}, complemented by openweb-math, algebraic stack, pes2o, arxiv, and StarCoder drawn from olmo-mix-1124~\citep{olmo20242} to provide domain-specialized coverage in reasoning, scientific, and programming knowledge. The 1B model is trained solely on SlimPajama for a lightweight and efficient pretraining. For tokenization, we use the tokenizer trained by \citet{bai2023qwen}, which offers efficient subword segmentation and robust multilingual support. For the 9B upcycled model, we use data sampled from the Nemotron Pretraining Dataset~\citep{basant2025nvidia}. For each node, the training data $\mathcal{D}_i$ for different nodes is randomly sampled from the whole dataset. Please see \textbf{Appendix~\ref{appendix:data_details}} for more details.

\textbf{Evaluation Details.} We evaluate our model using the \texttt{lm-evaluation-harness} library~\citep{eval-harness} and report results on several commonsense reasoning benchmarks, including SIQA~\citep{sap2019socialiqa}, ARC (easy and challenging)~\citep{clark2018think}, SciQ~\citep{SciQ}, PIQA~\citep{Bisk2020}, OpenBookQA~\citep{OpenBookQA2018}, WinoGrande~\citep{sakaguchi2021winogrande}, LogiQA~\citep{ijcai2020p501} and BoolQ~\citep{clark2019boolq}. To assess general knowledge, we utilize MMLU~\citep{hendrycks2020measuring} and C-Eval~\citep{huang2023c}. Additional evaluation details are included in the \textbf{Appendix~\ref{appendix:evaluation_details}}.

\subsection{Main Results}
\label{sec:main_results}

\textbf{Memory Cost Comparison.} Figs.~\ref{fig:memory_and_comunication_comparison} (a) and (c) compare the training memory footprints of SPES, DiLoCo, and centralized training. Both centralized training and DiLoCo require each node to update the full set of model parameters, resulting in high memory consumption. For example, training a 2B model requires more than 50GB memory per GPU, making it infeasible to train on commonly available 48GB GPUs. Furthermore, decentralized methods like DiLoCo cannot effectively leverage sharded training strategy due to limited inter-node bandwidth, further restricting the maximum trainable model size. In contrast, SPES keeps per-GPU memory under 40GB for a 2B model on 16 nodes without any sharding strategy. SPES can be combined with intra-node sharding for additional memory savings, as illustrated in Fig.~\ref{fig:memory_and_comunication_comparison}(c). This efficiency arises from sparse training: each node updates only a subset of parameters, substantially reducing per-GPU memory.

\textbf{Communication Cost Comparison.} 
Figs.~\ref{fig:memory_and_comunication_comparison} (b) and (d) compare the communication overhead of different training schemes. In each communication round, both DiLoCo and centralized training require every node to upload the full set of model parameters. In contrast, SPES uploads only the parameters that are actually updated, \textit{i.e.}, the shared modules and assigned experts, thereby substantially reducing communication overhead and training memory consumption. For instance, when training a 7B  model on 4 nodes, SPES requires only 9.8GB data to be uploaded per node per round, compared to 28.6GB for DiLoCo and centralized training—a reduction of 65\% in uplink communication volume. This demonstrates the communication efficiency brought by the sparse training strategy of SPES.

\begin{table*}[!t]
\caption{\textbf{Performance with and without expert merging.}}
\label{tab:expert_merging}
\begin{center}
\small
\begin{tabular}{lccccccccc}
\toprule
Method & ARC(e) & ARC(c) & PIQA & SciQ & OBQA & BoolQ & SIQA & WinoGrande & Avg. \\
\midrule
w/o merging & 48.77 & 24.41 & 65.83 & 74.10 & 31.20 & 53.85 & 42.58 & \textbf{51.14} & 48.99 \\
\rowcolor{myblue}
w/ merging  & \textbf{52.11} & \textbf{26.42} & \textbf{66.27} & \textbf{76.50} & \textbf{31.80} & \textbf{60.95} & \textbf{42.73} & 50.83 & \textbf{50.95} \\
\bottomrule
\end{tabular}
\end{center}
\end{table*}

\textbf{Training Speed Comparison.} We compare the training throughput of SPES against its centralized training counterpart. For the centralized setting, we adopt hybrid FSDP and train on four nodes, each equipped with 8×NVIDIA A800 GPUs and interconnected via RDMA. Each node contains four Mellanox InfiniBand HDR adapters, with each port operating at 100 Gbps (2×HDR lanes). In this configuration, centralized training reaches 3.79k tokens/s per GPU. Under the SPES setting (see section~\ref{exp:setup} for details), throughput with $H=50$ achieves 3.67k tokens/s. Despite running on a weaker hardware environment without high‑bandwidth interconnects, \textbf{SPES achieves a comparable speed}. In addition, its throughput can be further improved by reducing the synchronization frequency, highlighting its scalability under resource‑constrained conditions.

\textbf{Comparison with Previous Training Paradigms.} We evaluate SPES against both centralized training and the decentralized baseline DiLoCo, using 1B models trained on 50B tokens. As shown in Table~\ref{tab:performance_comparison_3_frameworks}, SPES achieves competitive performance on multiple benchmarks. Fig.~\ref{fig:result_comparison_3_frameworks} presents performance trajectories during training. Although SPES exhibits a slightly slower initial learning curve, attributable to its sparse expert updates, it rapidly converges and ultimately matches or outperforms both baselines. Notably, SPES achieves this with substantially lower per-node GPU memory consumption and reduced synchronization bandwidth relative to centralized and decentralized alternatives. These results highlight that SPES provides a favorable trade-off between computational efficiency and model quality, enabling decentralized pretraining to attain competitiveness with large-scale centralized training under significantly lower resource budgets.

\textbf{Performance Comparison with Existing LLMs.} Finally, to examine the scaling potential, we compare our 2B and 7B models, which are trained with less than 500B tokens, with those open-source models of similar activation parameter scales and trained with less than 3T tokens. The results are shown in Table~\ref{tab:commonsense}. We also show the results of models trained with significantly more tokens for reference. We can see that across several commonsense reasoning benchmarks, both our 2B and 7B models achieve competitive results among selected models with comparable activated parameter scales. It is worth noting that SPES-2B was trained in a decentralized manner on only 16 weakly connected 48GB GPUs, yet it remains competitive with models such as MobiLLama and OpenELM, which rely on substantially larger datasets and centralized infrastructures. This highlights the effectiveness of SPES in achieving strong performance under constrained hardware budgets. Moreover, SPES-7B attains results comparable to MoE++, which employs more advanced MoE designs (e.g., zero-computation experts) and larger training corpora. These findings indicate that SPES not only scales effectively and efficiently, but also retains significant room for improvement in architecture and data utilization, underscoring its potential as an extensible alternative to existing LLM training frameworks. 

Using a strong dense model as initialization, our largest model, SPES-9B achieves performance competitive with state-of-the-art models of comparable size using fewer than 500B tokens. We terminated training early due to resource constraints; however, metrics were still improving at the stopping point, indicating clear room for improvement.


\textbf{Expert-Merging Warm-Up.} As shown in Table~\ref{tab:expert_merging}, utilizing expert merging increases the average score from 48.99 to 50.95, with notable improvements on BoolQ and ARC. This indicates that cross-node parameter sharing enhances token utilization and promotes faster knowledge establishment, thus improving generalization across a range of reasoning and comprehension tasks.

\textit{For ablation studies on key hyperparameters, including the merging factor $\alpha$, warm-up steps $T_{w}$, merging interval steps $T_{i}$, merging Top-$K$, local training steps $H$, and the number of nodes $N$, please see \textbf{Appendix~\ref{appendix:additional_results}} for details.}
\section{Conclusion}

We introduced SPES, a decentralized and memory-efficient pretraining paradigm for MoE-based LLMs. SPES assigned distinct subsets of experts to individual nodes and synchronized them, substantially reducing per-device memory usage and communication overhead compared to centralized and prior decentralized approaches. To improve token utilization per expert, we introduced an expert-merging warm-up strategy to accelerate convergence in early training stages. Empirical results on 2B- and 7B-parameter MoE LLMs showed that SPES enabled efficient pretraining across weakly connected, geographically distributed GPU clusters, while achieving performance on par with comparable centralized baselines, and successfully scaled to upcycle a 9B model. Beyond lowering infrastructure demands, SPES broadened access to large-scale pretraining and could support more inclusive participation in LLM research, facilitating further advances in decentralized and memory-efficient training of foundation models.

\textbf{Limitations and Future Work.} Constrained by computational resources, our evaluation is limited to a 9B parameter model trained on less than 500B tokens. Validating scalability to larger models and extended training durations remains a critical direction for future research. Additionally, while this work focuses on language understanding, future efforts will investigate the applicability of SPES to multimodal reasoning and generative tasks. Extending the framework to these domains will provide a more comprehensive assessment of its generalization capabilities and limitations.
\section*{Impact Statement}

This work aims to advance Machine Learning. While it has potential societal implications, we identify no specific negative consequences requiring discussion.

\bibliography{ref}
\bibliographystyle{arxiv}

\newpage
\appendix
\onecolumn

\titlelabel{\thetitle.  }

\renewcommand{\thefigure}{A\arabic{figure}}
\renewcommand{\thetable}{A\arabic{table}}
\setcounter{figure}{0}
\setcounter{table}{0}

\section*{Appendix}
We provide the following materials in this appendix:

\ref{sec:theory}.~\textbf{Theoretical Analysis}: the convergence analysis of SPES.

\ref{appendix:implementation_details}.~\textbf{Implementation Details}: more details of training hyper-parameters.

\ref{appendix:data_details}.~\textbf{Data Details}: dataset descriptions and sampling ratios.

\ref{appendix:evaluation_details}.~\textbf{Evaluation Details}: evaluation datasets and metrics.

\ref{appendix:additional_results}.~\textbf{Additional Results}: results on additional benchmarks and ablations on hyper-parameters.


\ref{appendix:llm_usage}.~\textbf{Declaration of LLM Assistance}: description of LLM usage in manuscript preparation.

\section{Theoretical Analysis of SPES}
\label{sec:theory}

We study the convergence of \textbf{SP}arse \textbf{E}xpert \textbf{S}ynchronization (\textbf{SPES}). SPES performs \emph{block-sparse} local updates: all nodes update the shared parameters, while each expert block is updated only by its owner node. We also model the \emph{expert-merging warm-up} as an additional (early-stage) mixing perturbation applied after synchronization.

\subsection{Problem Setup and Notation}
\label{sec:theory_setup}

We minimize
\begin{equation}
\min_{\bm{\theta}} F(\bm{\theta})
\;:=\; \frac{1}{N}\sum_{i=1}^{N} f_i(\bm{\theta}),
\qquad
f_i(\bm{\theta})
\;:=\; \mathbb{E}_{\xi\sim \mathcal{D}_i}\bigl[\ell(\bm{\theta};\xi)\bigr],
\end{equation}
where each data sample $\xi$ is drawn from the local distribution $\mathcal{D}_i$. Here, $\ell(\bm{\theta};\xi)$ denotes the per-sample loss. $\bm{\theta}=(\bm{\psi},\bm{\Phi})$ and $\bm{\Phi}=\{\bm{\phi}_j\}_{j=1}^M$ are the shared and expert parameters, respectively. Let $\{\mathcal{P}_i\}_{i=1}^N$ be a partition of $\{1,\ldots,M\}$.
Node $\eta_i$ owns experts in $\mathcal{P}_i$; denote $o(j)$ the unique owner of expert $j$.

\subsection{SPES Update Rule}
\label{sec:theory_algo}

Let $\bm{\theta}^{(t)}$ be the global model at the beginning of round $t$. Each node sets $\bm{\theta}_i^{(t,0)}=\bm{\theta}^{(t)}$ and runs $H$ local stochastic gradient decent steps with step size $\eta$:
\begin{equation}
\bm{\theta}_i^{(t,h+1)}=\bm{\theta}_i^{(t,h)}-\eta\,\bm{U}_i\,\bm{g}_i^{(t,h)},
\qquad h=0,\ldots,H-1,
\label{eq:local_update}
\end{equation}
where $\bm{U}_i$ is a block mask that keeps updates only on $(\bm{\psi},\{\bm{\phi}_j:j\in\mathcal{P}_i\})$, and $\|\bm{U}_i\bm{v}\|\le \|\bm{v}\|$ for any vector $\bm{v}$.

\paragraph{Sparse synchronization.}
After $H$ steps, the server averages shared parameters and assigns each expert from its owner:
\begin{align}
\bm{\psi}^{(t+1,\mathrm{pre})}
&:= \frac{1}{N}\sum_{i=1}^{N} \bm{\psi}_i^{(t,H)},
\label{eq:sync_shared}
\\
\bm{\phi}_j^{(t+1,\mathrm{pre})}
&:= \bm{\phi}^{(t,H)}_{j,o(j)} \qquad \forall j.
\label{eq:sync_expert_assign}
\end{align}
Let $\bm{\theta}^{(t+1,\mathrm{pre})}=(\bm{\psi}^{(t+1,\mathrm{pre})},\bm{\Phi}^{(t+1,\mathrm{pre})})$ denotes the pre-merging parameters.

\paragraph{Expert-merging warm-up.}
For $t<T_{\mathrm{w}}$, we apply the merging step (Section~\ref{sec:method_spes}):
\begin{equation}
\bm{\phi}_j^{(t+1)}
:= \bm{\phi}_j^{(t+1,\mathrm{pre})}
+ \alpha_t \cdot \frac{1}{K}\sum_{k\in \mathcal{Q}_j}\Bigl(\bm{\phi}_k^{(t+1,\mathrm{pre})}-\bm{\phi}_j^{(t+1,\mathrm{pre})}\Bigr),
\label{eq:merge_step}
\end{equation}
with $\alpha_t\in[0,1]$ (and $\alpha_t=0$ for $t\ge T_{\mathrm{w}}$). Shared parameters are unchanged: $\bm{\psi}^{(t+1)}=\bm{\psi}^{(t+1,\mathrm{pre})}$.
Define the merge displacement
\begin{equation}
\Delta_{\mathrm{w}}^{(t+1)}:=\bm{\Phi}^{(t+1)}-\bm{\Phi}^{(t+1,\mathrm{pre})}.
\label{eq:merge_displacement_def}
\end{equation}

\paragraph{Equivalent pre-merge update.}
Define the per-round averaged stochastic directions (before merging)
\begin{align}
\widehat{\bm{g}}_{\bm{\psi}}^{(t)}
&:= \frac{1}{H}\sum_{h=0}^{H-1}\frac{1}{N}\sum_{i=1}^{N}\bm{g}_{i,\bm{\psi}}^{(t,h)},
\label{eq:ghat_shared}
\\
\widehat{\bm{g}}_{\bm{\phi}_j}^{(t)}
&:= \frac{1}{H}\sum_{h=0}^{H-1}\bm{g}_{o(j),\bm{\phi}_j}^{(t,h)}.
\label{eq:ghat_expert}
\end{align}
Then
\begin{equation}
\bm{\theta}^{(t+1,\mathrm{pre})}=\bm{\theta}^{(t)}-\gamma\,\widehat{\bm{g}}^{(t)},
\qquad \gamma:=\eta H.
\label{eq:global_premerge_update}
\end{equation}

\subsection{Assumptions}
\label{sec:theory_assumptions}

\paragraph{Assumption 1 (Smoothness).}
Each $f_i$ is $L$-smooth:
\begin{equation}
\|\nabla f_i(\bm{x})-\nabla f_i(\bm{y})\|\le L\|\bm{x}-\bm{y}\|,\quad \forall \bm{x},\bm{y},\ \forall i.
\label{eq:smoothness}
\end{equation}

\paragraph{Assumption 2 (Stochastic gradients).}
For all $i,t,h$,
\begin{equation}
\mathbb{E}\!\left[\bm{g}_i^{(t,h)} \mid \bm{\theta}_i^{(t,h)}\right]=\nabla f_i\!\left(\bm{\theta}_i^{(t,h)}\right),
\label{eq:unbiased}
\end{equation}
and there exist $\sigma_{\psi}^2,\sigma_{\Phi}^2\ge 0$ such that
\begin{align}
\mathbb{E}\!\left[\left\|\bm{g}_{i,\bm{\psi}}^{(t,h)}-\nabla_{\bm{\psi}} f_i(\bm{\theta}_i^{(t,h)})\right\|^2 \,\middle|\, \bm{\theta}_i^{(t,h)}\right]
&\le \sigma_\psi^2,
\label{eq:var_shared}
\\
\mathbb{E}\!\left[\sum_{j\in \mathcal{P}_i}\left\|\bm{g}_{i,\bm{\phi}_j}^{(t,h)}-\nabla_{\bm{\phi}_j} f_i(\bm{\theta}_i^{(t,h)})\right\|^2 \,\middle|\, \bm{\theta}_i^{(t,h)}\right]
&\le \sigma_\Phi^2,
\label{eq:var_experts}
\end{align}
and $\mathbb{E}\|\bm{g}_i^{(t,h)}\|^2\le G^2$ for some $G>0$.
(The last bound is used to control local drift.)

\paragraph{Assumption 3 (Expert-gradient heterogeneity).}
There exists $\zeta_{\Phi}\ge 0$ such that for all $\bm{\theta}$ and all $j$,
\begin{equation}
\left\|\nabla_{\bm{\phi}_j} f_{o(j)}(\bm{\theta})-\nabla_{\bm{\phi}_j} F(\bm{\theta})\right\|\le \zeta_{\Phi}.
\label{eq:heterogeneity_def}
\end{equation}
In particular, $\zeta_\Phi=0$ under IID data.

\paragraph{Assumption 4 (Bounded merge displacement).}
For $t<T_{\mathrm{w}}$, there exists $B_{\mathrm{w}}\ge 0$ such that
\begin{equation}
\mathbb{E}\!\left[\left\|\Delta_{\mathrm{w}}^{(t+1)}\right\|^2\right]\le \alpha_t^2\,B_{\mathrm{w}}^2.
\label{eq:merge_bounded}
\end{equation}

\subsection{Main Convergence Result}
\label{sec:theory_main}

\paragraph{Theorem 1 (Convergence of SPES).}
Suppose Assumptions 1--4 hold and $\gamma L\le \tfrac14$. Let $F_{\inf}:=\inf_{\bm{\theta}}F(\bm{\theta})$. Then for any $T\ge 1$, without expert warm-up merging

\begin{equation}
\begin{aligned}
\frac{1}{T}\sum_{t=0}^{T-1}\mathbb{E}\!\left[\|\nabla F(\bm{\theta}^{(t)})\|^2\right]
\;\le\;
&\;\frac{4\left(F(\bm{\theta}^{(0)})-F_{\inf}\right)}{\eta H\,T}
\;+\; 6\eta L\Bigl(\frac{\sigma_\psi^2}{N}+\sigma_\Phi^2\Bigr)
\;+\; 12 L^2\eta^2 H^2 G^2
\;+\; 12\,\zeta_\Phi^2
\label{eq:main_bound_premerge}
\end{aligned}
\end{equation}

Using expert warm-up merging,  we get

\begin{equation}
\begin{aligned}
\sup_{\bm{\theta}}\frac{1}{T}\sum_{t=0}^{T-1}\mathbb{E}\!\left[\|\nabla F(\bm{\theta}^{(t)})\|^2\right]
\;\propto\;
&\;\frac{4\left(F(\bm{\theta}^{(0)})-F_{\inf}\right)}{\eta H\,T}
\;+\; 6\eta L\Bigl(\frac{\sigma_\psi^2}{N}+\sigma_\Phi^2\Bigr)
\;+\; 12 L^2\eta^2 H^2 G^2
\;+\; 12\,\zeta_\Phi^2
\\
&\;+\;\frac{L\,B_{\mathrm{w}}^2}{\eta H\,T}\sum_{t=0}^{T_{\mathrm{w}}-1}\alpha_t^2 \leq Constant.
\label{eq:main_bound_merge_term}
\end{aligned}
\end{equation}

\paragraph{Discussion.}
The shared block enjoys $1/N$ variance reduction (term $\sigma_\psi^2/N$) due to averaging, while expert updates are owner-only (term $\sigma_\Phi^2$). The bias $\zeta_\Phi^2$ captures data heterogeneity in expert blocks. The merging warm-up appears as a vanishing perturbation when $\sum_{t<T_{\mathrm{w}}}\alpha_t^2$ is small (e.g., decaying $\alpha_t$ and $T_{\mathrm{w}}\ll T$).

\subsection{Proof of Theorem~1}
\label{sec:theory_proof}

We bound descent for the pre-merge iterate and then account for merging as a smooth perturbation.

\paragraph{1) Pre-merge descent.}
By $L$-smoothness, for $\bm{y}=\bm{x}-\gamma \bm{v}$,
\begin{equation}
F(\bm{y})\le F(\bm{x})-\gamma\langle \nabla F(\bm{x}),\bm{v}\rangle+\frac{L\gamma^2}{2}\|\bm{v}\|^2.
\label{eq:smooth_descent}
\end{equation}
Apply \eqref{eq:smooth_descent} with $\bm{x}=\bm{\theta}^{(t)}$, $\bm{v}=\widehat{\bm{g}}^{(t)}$, and write $\widehat{\bm{g}}^{(t)}=\nabla F(\bm{\theta}^{(t)})+\bm{e}^{(t)}$.
Using $\gamma L\le \tfrac14$ and AM–GM inequality yields
\begin{equation}
\mathbb{E}F(\bm{\theta}^{(t+1,\mathrm{pre})})
\le
\mathbb{E}F(\bm{\theta}^{(t)})
-\frac{\gamma}{4}\mathbb{E}\|\nabla F(\bm{\theta}^{(t)})\|^2
+\frac{3\gamma}{4}\mathbb{E}\|\bm{e}^{(t)}\|^2.
\label{eq:descent_premerge_4}
\end{equation}

\paragraph{2) Bounding the gradient error $\mathbb{E}\|\bm{e}^{(t)}\|^2$.}
Decompose $\bm{e}^{(t)}$ into variance (stochasticity) and bias (local drift + heterogeneity). Using Assumption~\eqref{eq:var_shared}--\eqref{eq:var_experts} and the averaging in \eqref{eq:ghat_shared}--\eqref{eq:ghat_expert} gives
\begin{equation}
\mathbb{E}\!\left[\left\|\widehat{\bm{g}}^{(t)}-\mathbb{E}[\widehat{\bm{g}}^{(t)}\mid \bm{\theta}^{(t)}]\right\|^2\right]
\;\le\; \frac{\sigma_\psi^2}{NH}+\frac{\sigma_\Phi^2}{H}.
\label{eq:variance_total}
\end{equation}
For the bias, local SGD drift over $h$ steps satisfies
$\mathbb{E}\|\bm{\theta}_i^{(t,h)}-\bm{\theta}^{(t)}\|^2\le \eta^2 h^2 G^2$
(from Assumption~2 and $\|\bm{U}_i\bm{v}\|\le\|\bm{v}\|$), hence by smoothness
$\mathbb{E}\|\nabla f_i(\bm{\theta}_i^{(t,h)})-\nabla f_i(\bm{\theta}^{(t)})\|^2\le L^2\eta^2 h^2 G^2$.
Averaging over $h\le H$ gives a bias contribution of order $L^2\eta^2 H^2 G^2$ on both shared and expert blocks, and Assumption~\eqref{eq:heterogeneity_def} adds $\zeta_\Phi^2$ on expert blocks. Overall,
\begin{equation}
\mathbb{E}\|\bm{e}^{(t)}\|^2
\;\le\;
2\Bigl(\frac{\sigma_\psi^2}{NH}+\frac{\sigma_\Phi^2}{H}\Bigr)
+4L^2\eta^2 H^2 G^2
+4\zeta_\Phi^2.
\label{eq:e_final_bound}
\end{equation}

\paragraph{3) Telescoping.}
Plug \eqref{eq:e_final_bound} into \eqref{eq:descent_premerge_4}, sum over $t=0,\ldots,T-1$, and use $F(\bm{\theta}^{(T,\mathrm{pre})})\ge F_{\inf}$ to obtain \eqref{eq:main_bound_premerge} (with $\gamma=\eta H$).

\paragraph{4) Effect of merging.}
Only experts change during merging, i.e., $\bm{\theta}^{(t+1)}-\bm{\theta}^{(t+1,\mathrm{pre})}=(\bm{0},\Delta_{\mathrm{w}}^{(t+1)})$.
By $L$-smoothness and Cauchy-Schwarz inequality,
\begin{equation}
F(\bm{\theta}^{(t+1)})
\le
F(\bm{\theta}^{(t+1,\mathrm{pre})})
+\frac{1}{2L}\|\nabla_{\bm{\Phi}}F(\bm{\theta}^{(t+1,\mathrm{pre})})\|^2
+L\|\Delta_{\mathrm{w}}^{(t+1)}\|^2.
\label{eq:merge_effect}
\end{equation}
Summing \eqref{eq:merge_effect} across rounds contributes an additive term proportional to
$\sum_t \mathbb{E}\|\Delta_{\mathrm{w}}^{(t+1)}\|^2$, yielding \eqref{eq:main_bound_merge_term} from Assumption~\eqref{eq:merge_bounded}.

\begin{table}[t]
\caption{\textbf{Training hyperparameters for different model scales.}}
\label{tab:hyperparams}
\centering
\begin{tabular}{lcccc}
\toprule
 & \textbf{9B} & \textbf{7B} & \textbf{2B} & \textbf{1B} \\
\midrule
Maximum Learning Rate    & $1\times10^{-4}$  & $4\times10^{-4}$ & $5\times10^{-4}$ & $5\times10^{-4}$ \\
Minimum Learning Rate    & $1\times10^{-5}$  & $4\times10^{-5}$ & $5\times10^{-5}$ & $5\times10^{-5}$ \\
Optimizer $\epsilon$     & $1\times10^{-8}$ & $1\times10^{-8}$ & $1\times10^{-8}$ & $1\times10^{-8}$ \\
Weight Decay             & 0.1  & 0.1              & 0.1              & 0.1 \\
$(\beta_{0}, \beta_{1})$ & (0.9, 0.95)  & (0.9, 0.95)      & (0.9, 0.95)      & (0.9, 0.95) \\
LR Warmup Steps          & 2000  & 2000             & 2000             & 2000 \\
Sequence Length          & 4096  & 2048             & 2048             & 2048 \\
Batch Size (Tokens)      & $0.5$M $\times 4$  & $2$M $\times 4$ & $0.5$M $\times 16$  & $0.5$M $\times 4$ \\
Synchronization Steps $H$  & 100 & 100              & 100              & 50 \\ 
\bottomrule
\end{tabular}
\end{table}

\section{Implementation Details}
\label{appendix:implementation_details}
Table~\ref{tab:hyperparams} details the full training configurations. For the from-scratch experiments (2B and 7B models), we adhere to these settings for the first 70\% of total training tokens; thereafter, we halve the per-node batch size and set $H=50$ to accelerate convergence. For the 1B model, the training budget is set to 50B tokens across all experiments, except for the ablations on $H$ and $N$, for which the budget is set to 100B tokens. For all experiments, the loss coefficients are fixed across the models as follows: cross-entropy ($1$), load-balancing ($0.01$), MoE z-loss ($0.001$), and standard z-loss ($1 \times 10^{-5}$).

\section{Details of Datasets and Sampling Ratio}
\label{appendix:data_details}

We train the model on data sampled from several open-source corpora, with sampling ratios provided in Table~\ref{tab:dataset_sample_ratio} and Table~\ref{tab:dataset_sample_ratio_upcycling}. Following~\citet{olmo20242}, we apply a filter that removes all documents containing sequences of 32 or more repeated $n$-grams (an $n$-gram denotes any span of 1–13 tokens). The uses datasets are summarized as follows.

\textbf{Ultra-FineWeb.} Ultra-FineWeb~\citep{wang2025ultra} is a large-scale web corpus constructed from FineWeb~\citep{penedo2024fineweb} and Chinese FineWeb~\citep{yu2025opencsg} using an efficient verification-based filtering pipeline. The approach combines lightweight fastText classification with a verification mechanism, enabling reliable data selection at substantially reduced computational cost. The final corpus comprises roughly 1 trillion English tokens and 120 billion Chinese tokens. By enhancing overall data quality, Ultra-FineWeb provides a strong foundation for LLM training and contributes to the dataset used in MiniCPM4~\citep{team2025minicpm4}.

\textbf{OLMo-Mix-1124.}  OLMo-Mix-1124 is a 3.9-trillion-token corpus comprising over 95\% web data, constructed from DCLM~\citep{li2024datacomp}, Dolma v1.7~\citep{soldaini2024dolma}, and StarCoder~\citep{lozhkov2024starcoder}. For our work, we extract scientific-domain subsets, including arXiv, OpenWebMath, Algebraic Stack, peS2o, and StarCoder.

\textbf{Nemotron Pretraining Dataset.\footnote{\url{https://huggingface.co/collections/nvidia/nemotron-pre-training-datasets}}} Nemotron-Pretraining~\citep{basant2025nvidia} is a large-scale corpus collected for the NVIDIA Nemotron Nano 2 family, this dataset emphasizes high-value math, code, and multilingual Q\&A to fuel globally-capable models. It aggregates four specialized components: a 133B-token math corpus~\citep{mahabadi2025nemotron} processed via a novel Lynx + LLM pipeline, an updated English web crawl enriched with synthetic data~\citep{su2025nemotron}, a rigorously filtered source code dataset, and a diverse SFT-style collection covering STEM and reasoning domains.

\textbf{SlimPajama.} SlimPajama~\citep{cerebras2023slimpajama} is a large-scale, rigorously deduplicated corpus constructed from RedPajama~\citep{weber2024redpajama}. Using a multi-stage pipeline that combines quality filtering with MinHashLSH-based deduplication at trillion-token scale, SlimPajama substantially reduces redundancy and low-quality content, compressing the dataset from 1.21T to 627B tokens while retaining domain coverage. The corpus spans diverse sources, including CommonCrawl, C4, GitHub, Books, ArXiv, Wikipedia, and StackExchange.

\begin{table}[t]
    \caption{\textbf{Dataset sampling ratios for the from-scratch training regimen.}}
    \label{tab:dataset_sample_ratio}
    \small
    \centering
    \setlength{\tabcolsep}{5pt}
    \begin{tabular}{lccccccc}
        \toprule
        \textbf{Dataset} 
            & Ultra-FineWeb 
            & SlimPajama 
            & StarCoder 
            & arXiv 
            & OpenWebMath 
            & Pes2o 
            & Algebraic Stack \\
        \midrule
        \textbf{Ratio (\%)} 
            & 64.2 
            & 27.2 
            & 6.6  
            & 0.7  
            & 0.4  
            & 0.5  
            & 0.4  \\
        \bottomrule
    \end{tabular}
\end{table}

\begin{table}[t]
    \caption{\textbf{Dataset sampling ratios for the upcycling training regimen.}}
    \label{tab:dataset_sample_ratio_upcycling}
    \small
    \centering
    \setlength{\tabcolsep}{5pt}
    \begin{tabular}{lcccc}
        \toprule
        \textbf{Dataset} 
            & Nemotron-CC-V2 
            & Nemotron-Math-V1 
            & Nemotron-Pretraining-Code 
            & Nemotron-Pretraining-SFT  \\
        \midrule
        \textbf{Ratio (\%)} 
            & 63.3 
            & 16.4 
            & 11.9  
            & 8.4   \\
        \bottomrule
    \end{tabular}
\end{table}

\begin{table}[t]
  \caption{\textbf{Performance comparison with previous LLMs on additional benchmarks.} Some models are excluded because they neither report results on these benchmarks nor are compatible with \texttt{lm-evaluation-harness}. }
  \label{tab:main_results_knowledge}
  
  \small
  \renewcommand{\arraystretch}{1.1}
  \begin{center}
  \begin{tabular}{l
            >{\centering\arraybackslash}p{1.2cm}
            >{\centering\arraybackslash}p{1.1cm}
            >{\centering\arraybackslash}p{1.1cm}
            >{\centering\arraybackslash}p{1.1cm}
            >{\centering\arraybackslash}p{1.1cm}  
            >{\centering\arraybackslash}p{1.1cm}  
    }
    \toprule
    \textbf{Method} & \textbf{\#Params} & \textbf{\#Tokens} & \textbf{OBQA} & \textbf{LogiQA} & \textbf{C-Eval} & \textbf{MMLU} \\
    \midrule
    \multicolumn{6}{c}{\textbf{Models Trained with Significantly More Tokens}} \\
    \midrule
    \rowcolor{lightgray}
    Qwen2.5-0.5B~\citep{qwen2025qwen25technicalreport} & 0.5B/0.5B & 18T & 35.4 & 29.5 & 51.0 & 47.3 \\
    \rowcolor{lightgray}
    Qwen3-0.6B~\citep{yang2025qwen3} & 0.6B/0.6B & 36T & 34.2 & 29.0 & 50.4 & 52.8 \\
    \rowcolor{lightgray}
    Llama3.2-1B~\citep{dubey2024llama} & 1.1B/1.1B & 9T & 36.2 & - & 30.9 & 36.6 \\
    \rowcolor{lightgray}
    Qwen2.5-1.5B~\citep{qwen2025qwen25technicalreport} & 1.5B/1.5B & 18T & 40.4 & 31.5 & 68.2 & 59.7 \\
    \rowcolor{lightgray}
    SmolLM2-1.7B~\citep{allal2025smollm2} & 1.7B/1.7B & 11T & 43.6 & 29.8 & 32.5 & 48.4 \\
    \rowcolor{lightgray}
    Qwen3-1.7B~\citep{yang2025qwen3} & 1.7B/1.7B & 36T & 38.6 & 31.5 & 65.2 & 62.6 \\
    \rowcolor{lightgray}
    OLMoE-1B-7B~\citep{muennighoff2024olmoe} & 1.3B/7B & 5T & 45.2 & 28.4 & 31.1 & 50.5 \\
    
    \midrule
    \multicolumn{6}{c}{\textbf{Models with $\leq 3$B Parameters}} \\
    \midrule
    MobiLlama-0.8B~\citep{thawakar2024mobillama} & 0.8B/0.8B & 1.3T & 33.0 & - & 22.7 & 23.5 \\
    TinyLlama-1.1B~\citep{zhang2024tinyllama} & 1.1B/1.1B & 3T & 36.8 & 26.3 & 26.0 & 25.3 \\
    OPT-1.3B~\citep{zhang2022opt}  & 1.3B/1.3B & 180B & 33.4 & 26.9 & 23.0 & 24.9 \\
    MobiLlama-1.3B~\citep{thawakar2024mobillama} & 1.3B/1.3B & 1.3T & 35.4 & - & 26.2 & 25.3 \\
    Pythia-1.4B~\citep{biderman2023pythia}   & 1.4B/1.4B & 300B & 33.4 & 27.3 & 23.0 & 24.2 \\
    OPT-2.7B~\citep{zhang2022opt} & 2.7B/2.7B & 180B & 35.2 & 26.0 & 23.0 & 25.6 \\
    Pythia-2.8B~\citep{biderman2023pythia}  & 2.8B/2.8B & 300B & 35.6 & 28.0 & 22.9 & 25.2 \\
    
    \rowcolor{myblue}
    \textbf{SPES-2B (ours)} & 0.8B/2.1B & 500B & 31.4 & 27.2 & 25.0 & 25.5 \\

    \midrule    
    \multicolumn{6}{c}{\textbf{Models with 7B Parameters}} \\
    \midrule
    MoE++ 7B~\citep{jin2024moe++} & 1.2B/7B  & 1T & 40.0 & 28.4 & 23.6 & 24.6 \\
    LLaMA-MoE-3.0B~\citep{llama-moe} & 3.0B/7B & 2.2T & - & 30.6 & - & 26.8 \\
    \rowcolor{myblue}
    \textbf{SPES-7B (ours)} & 1.6B/7B & 500B & 39.4 & 27.5 & 26.2 & 24.9 \\
    \rowcolor{myblue}
    \textbf{SPES-9B (ours)} & 3.1B/9B & 400B & 42.2 & 30.4 & 44.7 & 63.7 \\
    \bottomrule
  \end{tabular}
  \end{center}
\end{table}

\begin{table}[!t]
\caption{\textbf{Performance comparison with different numbers of nodes.}}
\label{tab:performance_num_nodes}
\centering
\small
\begin{tabular}{cccccccccc}
\toprule
No. of Nodes & ARC(e) & ARC(c) & PIQA & SciQ & OBQA & BoolQ & SIQA & WinoGrande & Avg. \\
\midrule
2 & 52.0 & 25.7 & 68.7 & 77.6 & 30.4 & 58.0 & 42.2 & 50.4 & 50.6 \\
4 & 51.8 & 27.4 & 67.4 & 75.3 & 29.8 & 49.5 & 43.7 & 52.6 & 49.7 \\
8 & 47.9 & 24.6 & 66.3 & 70.8 & 29.4 & 60.1 & 42.8 & 53.9 & 49.5 \\
\bottomrule
\end{tabular}
\end{table}

\section{Evaluation Details}
\label{appendix:evaluation_details}

We evaluate our models with the \texttt{lm-evaluation-harness} library~\citep{eval-harness}, which offers standardized benchmark implementations and facilitates direct comparison with prior work. All experiments use version 0.4.7. The benchmarks and evaluation settings are detailed below:

\textbf{SciQ}~\citep{SciQ} is a science multiple-choice question-answering dataset. The questions were generated by crowdworkers and validated against science reference materials, covering topics such as physics, biology, and chemistry. As the questions are designed to resemble real exam-style queries, the dataset tests scientific knowledge and reasoning skills of a model. We report 0-shot accuracy on SciQ.

\textbf{ARC}~\citep{clark2018think} (AI2 Reasoning Challenge) consists of grade-school level science exam questions, partitioned into ARC-Easy (ARC-E) and ARC-Challenge (ARC-C). ARC-E contains questions that can often be answered by retrieval of surface-level facts, while ARC-C includes the more demanding questions requiring reasoning and multi-step inference across scientific facts. We report 0-shot accuracy on ARC-E and 25-shot normalized accuracy on ARC-C.

\textbf{SIQA}~\citep{sap2019socialiqa} (SocialIQA) benchmarks social commonsense reasoning. Each instance presents a short human-centered scenario alongside a question about likely intents, causes, or outcomes of human actions. This evaluates the model’s ability to handle subtle social reasoning and cause-effect relationships in naturalistic settings. We report 0-shot normalized accuracy on SIQA.

\textbf{PIQA}~\citep{Bisk2020} (Physical Interaction QA) evaluates physical commonsense reasoning in everyday situations. Given a description of a goal, the model must choose the most plausible solution among two alternatives, testing physical feasibility and everyday world knowledge. We report 0-shot normalized accuracy on PIQA.

\textbf{OpenBookQA}~\citep{OpenBookQA2018} presents multiple-choice science questions paired with a small open-book of 1,326 core scientific facts. Answering the questions typically requires combining knowledge from the book with additional commonsense reasoning, making this benchmark particularly challenging. We report 0-shot normalized accuracy.

\textbf{WinoGrande}~\citep{sakaguchi2021winogrande} is a large-scale dataset for pronoun resolution, created to reduce annotation artifacts common in earlier benchmarks (e.g., Winograd Schema Challenge). Each instance requires the model to resolve ambiguous pronouns based on contextual clues, testing commonsense reasoning and language understanding. We report 0-shot accuracy on WinoGrande.

\textbf{BoolQ}~\citep{clark2019boolq} is a reading comprehension dataset in the yes/no QA format. Questions are naturally occurring user queries, paired with passages from Wikipedia that may or may not contain the answer. Models must perform passage-level understanding to correctly infer the response. We report 0-shot accuracy on BoolQ.

\textbf{C-Eval}~\citep{huang2023c} is a comprehensive Chinese evaluation suite consisting of over 13,000 multiple-choice questions spanning 52 subjects, from elementary school topics to professional certification exams. It provides a fine-grained view of model performance in academic and professional domains under Chinese cultural and linguistic settings. We report 0-shot accuracy on C-Eval. 

\textbf{LogiQA}~\citep{ijcai2020p501} is a dataset sourced from expert-written questions designed to evaluate machine reading comprehension through logical reasoning. It consists of 8,678 QA instances that cover multiple types of deductive reasoning, serving as a benchmark where state-of-the-art models still trail the human ceiling. We report 0-shot normalized accuracy.

\textbf{MMLU}~\citep{hendrycks2020measuring} (Massive Multitask Language Understanding) covers 57 tasks across diverse domains such as mathematics, history, law, medicine, and the natural sciences. As a broad knowledge benchmark, it measures both factual recall and domain-specific reasoning. We follow standard settings and report 5-shot accuracy on MMLU.

\begin{figure}[t]
  \centering
  \includegraphics[width=\linewidth]{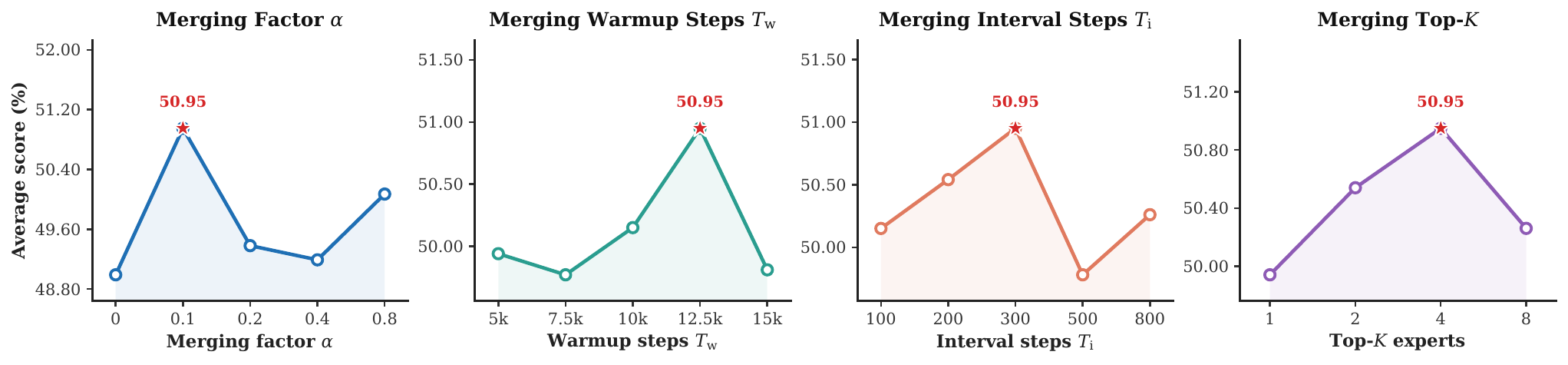}
  \caption{\textbf{Ablation on key hyper-parameters in expert merging.} The reported average is computed over ARC(e), SciQ, PIQA, WinoGrande, ARC(c), OBQA, OpenBookQA, and SIQA.}
  \label{fig:ablation_alpha_tmerge}
\end{figure}

\begin{figure}[t]
  \centering
  \includegraphics[width=\linewidth]{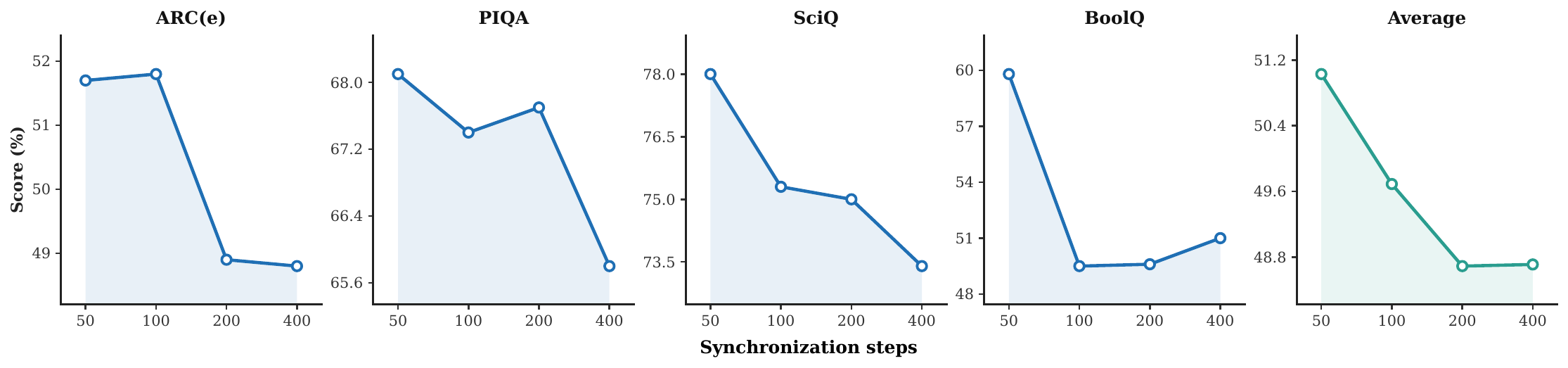}
  \caption{\textbf{Ablation on synchronization steps.} The reported average is computed over eight benchmarks in total, additionally including ARC(c), OBQA, OpenBookQA, and SIQA.}
  \label{fig:different_sync_steps}
\end{figure}

\section{Additional Results}
\label{appendix:additional_results}

\textbf{Results on Additional Benchmarks.} Table~\ref{tab:main_results_knowledge} reports the performance of our models on additional benchmarks. On general knowledge benchmark, SPES-7B surpasses the comparable baseline MoE++ (26.2 vs. 23.6 on C-Eval, 24.9 vs. 24.6 on MMLU), while maintaining competitive performance on other tasks. This indicates that SPES can match the performance of centrally trained models under resource-constrained settings, underscoring its potential to lower the barrier to LLM pretraining. In addition, SPES-2B attains performance on par with models of similar scale using only 16 weakly connected nodes, further validating our approach.

\textbf{Ablation on Number of Nodes.} We then study the impact of varying the number of nodes $N$ while keeping the global batch size fixed. As shown in Table~\ref{tab:performance_num_nodes}, model performance remains stable when scaling from 2 to 8 nodes. The average score decreases slightly from 50.6 (2 nodes) to 49.5 (8 nodes), yet SPES maintains competitive results across benchmarks. This behavior illustrates a natural trade-off in decentralized sparse training: increasing the number of nodes leads to greater fragmentation of training data and experts, which can modestly slow convergence. Nonetheless, the results underscore the robustness of SPES. Even with reduced per-node token utilization, it maintains overall performance. These findings demonstrate SPES' potential of scalability, suggesting that it can effectively leverage a larger number of participants while maintaining model quality, a key property for practical deployment in heterogeneous, distributed environments.

\textbf{Ablation on Hyperparameters in Expert Merging.} Fig.~\ref{fig:ablation_alpha_tmerge} shows the effect of varying merging warmup steps $T_{w}$, merging interval steps $T_{i}$, merging factor $\alpha$ and merging Top-$K$ on performance. A moderate warmup of 12.5k steps achieves the best results, as shorter schedules hinder sufficient knowledge exchange, while excessively long ones interfere with expert specialization. For the merging interval, performance peaks at $T_i=300$, while both more frequent and less frequent merging lead to lower performance. Similarly, performance peaks when $\alpha$ is set to $0.1$ and $K$ is set to $4$, with both smaller and larger values leading to degradation. These observations suggest that effective expert merging requires a careful balance between inter-expert knowledge sharing and expert specialization. Overly aggressive merging may overwrite expert-specific information, whereas insufficient merging yields only minor parameter updates and limits the efficiency of knowledge sharing, thereby slowing the establishment of general expert representations.

\textbf{Ablation on Synchronization Steps.} We analyze the effect of varying the local update interval $H$ in the SPES framework. As illustrated in Fig.~\ref{fig:different_sync_steps}, performance declines when $H$ increases from 50 to 200 or 400. This trend reflects a key trade-off in decentralized sparse training: while larger $H$ reduces communication frequency, it amplifies model divergence across nodes, weakening the benefits of expert sharing.  Overall, $H=50$ provides the best balance between communication efficiency and model quality, underscoring the necessity of frequent synchronization to fully exploit SPES' sparse expert updates under bandwidth-limited decentralized settings.


\section{Declaration of LLM Assistance}
\label{appendix:llm_usage}

We use ChatGPT‑5 to assist with the refinement of this manuscript. After drafting the full text, we provided selected passages to the models for suggestions on grammar, clarity, and conciseness. All revisions were reviewed and finalized by the authors to ensure accuracy and appropriateness.

\end{document}